\documentclass[letterpaper, 10 pt, conference]{ieeeconf}
\usepackage{graphicx} 
\usepackage{hyperref}
\usepackage{subcaption}
\usepackage{colonequals}
\usepackage{amsmath} 
\usepackage{float}
\usepackage{booktabs}
\usepackage{amsfonts}
\usepackage{amssymb}
\usepackage[dvipsnames]{xcolor}
\usepackage{siunitx}
\usepackage{ulem}
\usepackage[font=small,labelfont=bf]{caption}
\DeclareMathOperator*{\argmax}{arg\,max}

\usepackage[ruled]{algorithm2e}
\usepackage{multirow}
\makeatletter
\newcommand{\algorithmstyle}[1]{\renewcommand{\algocf@style}{#1}}
\newcommand{\removelatexerror}{\let\@latex@error\@gobble}
\newcommand{\norm}[1]{\left\lVert#1\right\rVert}
\makeatother
\usepackage[noadjust]{cite}

\IEEEoverridecommandlockouts                              

\title{\LARGE \bf
	Generalizable task representation learning from human demonstration videos: a geometric approach
}

\author{Jun Jin$^{\dagger}$ and Martin Jagersand$^{\dagger}$
\thanks{$^{\dagger}$Authors are with the Department of Computing Science,
        University of Alberta, Edmonton AB., Canada, T6G 2E8.
        { 
           \tt\small \{jjin5,mj7\} @ualberta.ca
        }
        }%
}

\begin{document}
	\maketitle
	\thispagestyle{empty}
	\pagestyle{empty}
	
	\begin{abstract}
We study the problem of generalizable task learning from human demonstration videos without extra training on the robot or pre-recorded robot motions. Given a set of human demonstration videos showing a task with different objects/tools (categorical objects), we aim to learn a representation of visual observation that generalizes to categorical objects and enables efficient controller design. We propose to introduce a geometric task structure to the representation learning problem that geometrically encodes the task specification from human demonstration videos, and that enables generalization by building \textit{task specification correspondence} between categorical objects. Specifically, we propose CoVGS-IL, which uses a graph-structured task function to learn task representations under structural constraints. Our method enables task generalization by selecting geometric features from different objects whose inner connection relationships define the same task in geometric constraints. The learned task representation is then transferred to a robot controller using uncalibrated visual servoing (UVS); thus, the need for extra robot training or pre-recorded robot motions is removed.
	\end{abstract}
	
	\section{Introduction}
	\label{sec:intro}
One interesting phenomenon in human learning by watching demonstrations is that we can not only infer the task definition (specification) from one demonstration video, but also generalize the same task specification by watching a collection of demonstration videos showing the same task using various objects or tools. In the low level, each video in the collection individually encodes a realization of the task with a specific object or tool demonstrated. In the high level, each video encodes a task specification that is the same to each other since the same task is defined. Therefore, there is a \textbf{\textit{task specification correspondence}} between each demonstration video of the collection since the same task specification builds correspondence between task settings using different objects. Can we apply the same learning mechanism in robotics to enable generalizable task learning?

This problem is challenging in two folds. (1) First, if we put aside task generalization, the problem of robot learning by watching human demonstrations~\cite{stadie2017third,sharma2019third,xiong2016robot,paulius2016functional, billard2004robot, Argall2009} is difficult since we only have video samples without any clues about robot actions. Enabling a robot to perform the task will require either tedious robot training~\cite{stadie2017third,sharma2019third} or pre-recorded motions~\cite{xiong2016robot,paulius2016functional}, which impede its viability in the real world. (2) Learning generalizable skills from a collection of human demonstration videos remains an unsolved problem. Questions regarding what should be learned to achieve task generalization and why the learned model is generalizable lack sufficient research attention. For example, should we enable generalization in the representation learning level or the policy learning level that maps observation to robot actions? Furthermore, suppose we can learn a generalizable representation or policy using deep neural networks, it remains challenging to interpret what has been learned to enable task generalization. 

To address the above problems, we introduce a geometric task structure to learn generalizable representations from human demonstration videos. Commonly, there are two types of task structures: (1) a geometric task structure that uses geometric features (points, lines, planes)   to compose projective-invariant constraints (point-to-point, parallel lines) to define a task; (2) a semantic task structure~\cite{saxena2014robobrain} that extracts task semantic meanings in the form of knowledge graphs~\cite{paulius2019survey}, grammar trees~\cite{yang2015robot}, behaviour trees~\cite{paxton2017costar}  or the planning domain definition language (PDDL)~\cite{leidner2012things}. We choose a geometric task structure as the inductive bias in learning since its adjoint outputs---geometric errors, compared to a semantic task structure’s output, are more friendly to robot controllers as shown in the visual servoing literature~\cite{Chaumette2006}. As a result, it is likely to remove the requirement for tedious robot training or pre-recorded motions.

\begin{figure}[tbp]
	\setlength{\belowcaptionskip}{-10pt}
		\begin{center}
			\includegraphics[width=0.5\textwidth]{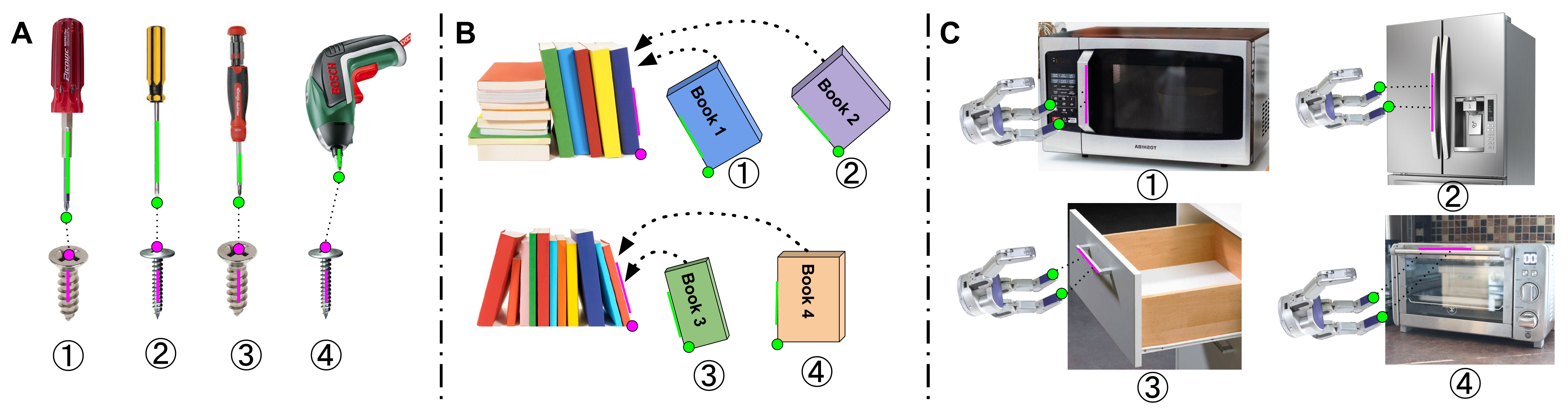} 
			\caption{Task specification correspondence and examples of using geometric task structures to specify a manipulation task (\textbf{A:} a screwing task. \textbf{B:} a book organization task. \textbf{C:} a door, drawer open task). For different objects/tools, the same type of geometric constraints (such as in Fig. 1A, point-to-point and line-to-line) defines the same task (screwing), which we call as \textit{task specification correspondence} since the same task definition builds correspondence between task settings using different objects. Can we use this clue for generalizable task learning?
			}
			\label{fig:design_overview}
		\end{center}
	\end{figure}
		\begin{figure*}[h]
	\setlength{\belowcaptionskip}{-10pt}
	\begin{center}
		\includegraphics[width=0.9\textwidth]{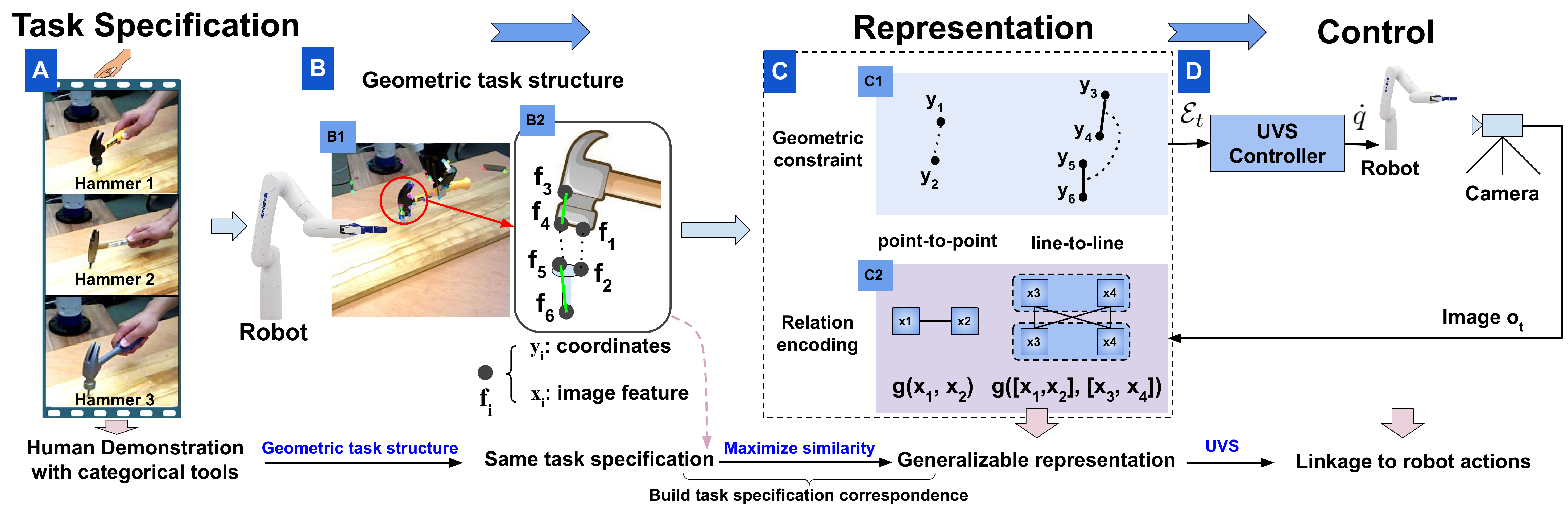} 
		\caption[Overview of our method]{\small
			Overview of our method explained using a hammering task. Given human demonstration videos (Fig. {A}) showing different types of hammers, we aim to learn a representation that generalizes to different hammers by introducing a geometric task structure (Fig. {B}) which constrains the representation space using predefined graph structures (Fig. C2), where each structure defines a geometric constraint type. Task generalization is achieved by maximizing a similarity metric between the representation of geometric constraints from different hammers. The learned representation is then mapped to robot actions (Fig. D) by its adjoint geometric error output $\mathcal{E}_{t}$ that links to an uncalibrated visual servoing (UVS) controller~\cite{Jagersand1997}. During testing, given an image observation $o_{t}$, the learned representation selects image features to fit in the graph nodes and outputs geometric errors $\mathcal{E}_{t}$ for robot control. In this example, $\mathcal{E}_{t}$ is computed from the coordinate difference between two points, $[y_{2} - y_{1}]$ and the parallelism metric between two lines, $[(y_{3} \times y_{4}) \times (y_{5} \times y_{6})]$.
			
		}
		\label{big_overview}
	\end{center}
\end{figure*}
We show that introducing a geometric structure in learning also brings task generalization by building \textit{task-specification correspondence} (Fig. 1), which means the task definition stays consistent under various task settings, such as different tools, objects or backgrounds. As a result, the consistency of task specification builds correspondence between different task settings, which is the key to achieve task generalization. 

Specifically, we show that a geometric task structure can be used to build \textit{task-specification correspondence} since the same type of geometric constraints defines the same task by using image features on different objects. From a representation perspective, a geometric constraint is a connection relationship between geometric features (as further described in Fig. 3 and 4), which can be parameterized using relation encoders (Fig. 2 C2), such as graph neural networks (GNNs~\cite{wu2020comprehensive}) or visual transformers~\cite{han2020survey}. Thus, maximizing a similarity metric between representations of geometric constraints from different objects/tools will build the \textit{task-specification correspondence}. This concludes the basic idea of our proposed method. To summarize, our contributions are as follows.
\begin{itemize}
    \item We introduce a geometric task structure to learn generalizable task representations from human demonstration videos. Our proposed method learns to select task-relevant geometric constraints on categorical objects with the same task functionality to build \textit{task specification correspondence} which enables generalization.
    \item We propose a novel approach to robot learning by watching human demonstrations. Our approach tackles visual imitation learning from a geometric perspective and combines it with conventional visual servoing that removes the need for extra training on the robot or pre-recorded robot motions.
  \end{itemize}
We use a hammering task running on a Kinova Gen3 robot as an example  with comparison to various baselines to support our contributions.

\section{Related works}

\subsection{Geometric task structure in robotics}

Humans have been using points and lines to describe structural concepts for thousands of years. Many tasks in our everyday life can be described using geometric features. Likewise, in robotics, composing a task from image or point cloud data by the association, combinations and sequential linkage of geometric features is studied in the visual servoing literature, including the theoretical frameworks~\cite{hager2000specifying,chaumette1994visual, hespanha1999tasks}, system ~\cite{Dodds1999} and applications ~\cite{Gridseth2016}.

In addition to traditional methods, geometric task structures are also used as an intermediate representation in robot learning to improve sample efficiency and task generalization. For example, S. Levine et al. 2015~\cite{levine2016end} and their following works~\cite{finn2016guided} report sample efficiency using the spatial-softmax operator to enforce the neural network extracting task-relevant feature point structures. Qin et al. 2020~\cite{qin2020keto} show task generalization in reinforcement learning by representing the task using keypoint structures. 

\textbf{Our work} extends key point-based approaches to a more general geometry-based approach. We give a more thorough discussion on using geometric features like points, lines and conics to represent a task. Moreover, we take extra considerations on studying how to design a proper parameterization of \textit{what} to make designing \textit{how} much easier.

\subsection{Generalizable visual descriptors}
This paper is also inspired by the various approaches that learn generalizable visual descriptors for robotic tasks. Commonly, these approaches learn generalizable visual descriptors based on 3D correspondence, involving 3D reconstruction and then reprojecting the generated correspondence samples on the images. For example, \textit{Dense Object Descriptor}~\cite{florence2018dense}, which learns a descriptor based on 3D reconstruction and image reprojection to build correspondence. Their further work kPAM~\cite{manuelli2019kpam} uses contrastive learning to explore learning dense object descriptor that is invariant across categorical objects. However, such training requires laborious annotations on each of the images. 

\textbf{Our work} proposes to build the correspondence in an unsupervised way by extracting task-relevant geometric constraints across categorial objects. In addition to the task generalization gain in our approach, we show that using geometric constraints is more controller friendly that can be directly applied to a geometric vision controller (visual servoing~\cite{chaumette1994visual}). As a comparison, the above methods require tediously hand-engineered robot controllers.


\section{Method}
\label{sec:method}
In this section, we firstly introduce the concept and parameterization of a geometric task structure. Then we explain how to achieve task generalization by building \textit{task specification correspondence}. At last, we explain how to learn a generalizable task representation from human demonstration videos and map the learned representation to robot actions.

\subsection{Geometric task structure and parameterization}
\begin{figure}[h]
\setlength{\belowcaptionskip}{-10pt}
	\setlength{\belowcaptionskip}{-10pt}
	\begin{center}
		\includegraphics[width=0.4\textwidth]{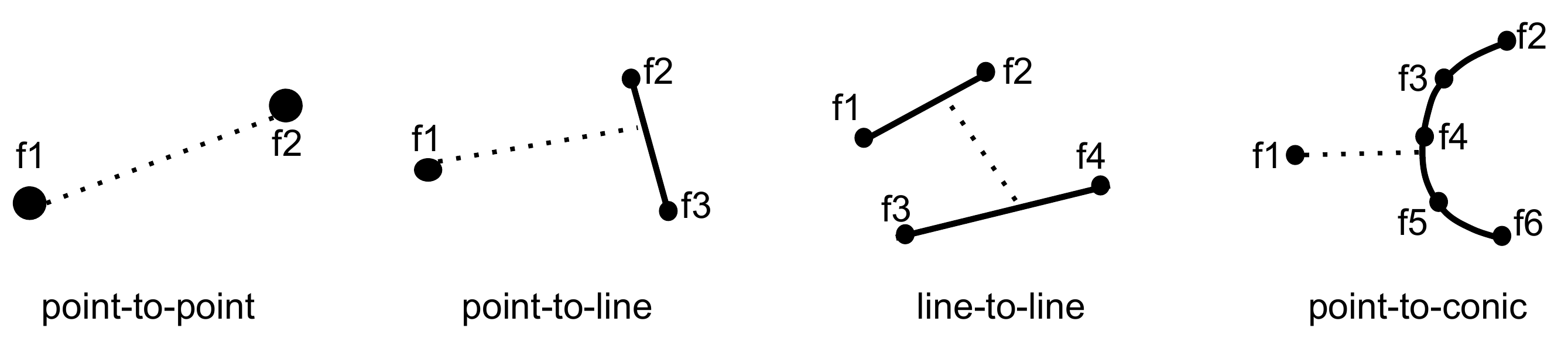} 
		\caption[Relationship of geometric features.]{ An arbitrary geometric constraint is a multi-entity relationship between a set of feature points, wherein the low level, the inner-connections (solid line) between feature points $\{f_{i}\}$ define a complex geometric feature and in the high level, the outer-connection (dashed line) between geometric features  defines different geometric constraints.}
		\label{fig:current}
	\end{center}
\end{figure}

Generally, an arbitrary geometric constraint is a binary relationship (Fig. \ref{fig:current} dashed lines) between two geometric primitives (point, line, conic and plane) observed in the image space. Considering the difficulty of representing complex geometric primitives using neural networks and that complex geometric features can be further decomposed to discrete points (Fig. \ref{fig:current} solid lines). A geometric constraint can be represented as a multi-entity relationship between a set of feature points $\{f_{i}\}$ (Fig. \ref{fig:current}).

The relationship is parameterized using an undirected graph $\mathcal{G}=\{V, E\}$, with each graph node $v \in V$ representing an image feature. The connections $E$ between graph nodes define (1) how complex geometric features are constructed from feature points (Fig. 3 solid lines); (2) how geometric features are associated as a geometric constraint (Fig. 3 dashed lines). 

Using a graph to encode the relationship between image features to represent a geometric constraint was proposed in our previous work~\cite{jin2020geometric,jin2020visual}, where the graph structure is derived by examining each possible node connection considering two properties~\cite{jin2020geometric}: permutation-invariant and non-inner-associative. For example, a line-to-line constraint has 38 possible node connections and the only one that fulfils the two properties is selected out (shown in Fig. 4). Details about how the graph structure is derived can be found in our supplementary material~\cite{proofs21}.

\begin{figure}[h]
\setlength{\belowcaptionskip}{-10pt}
	\includegraphics[width=0.35\textwidth]{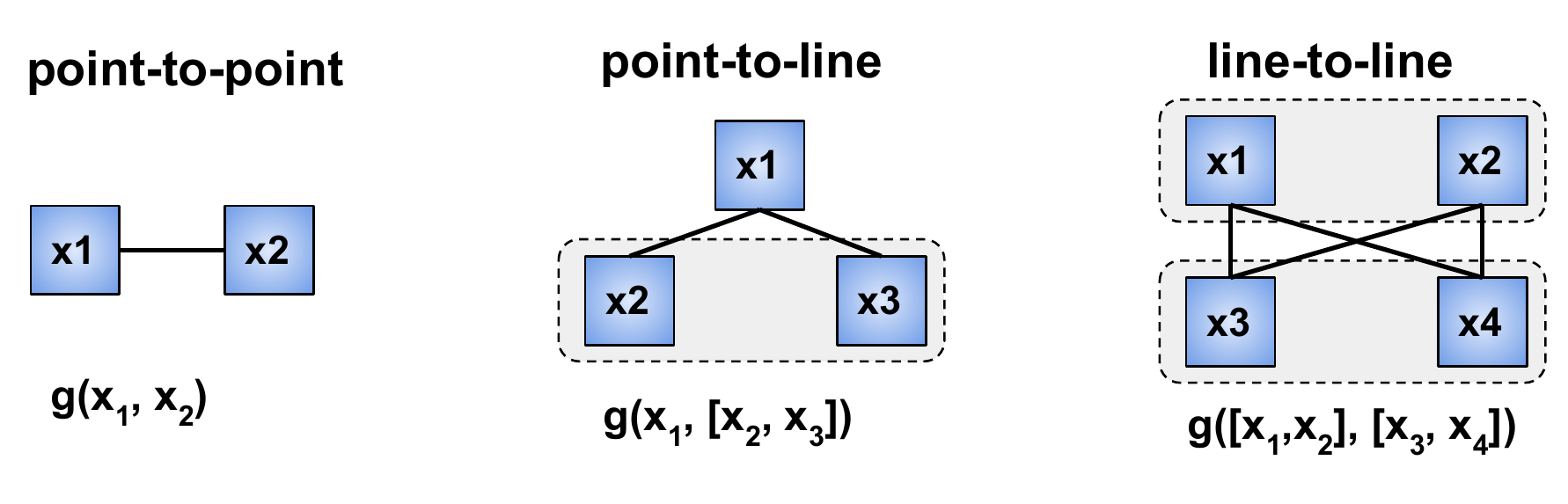}
	\centering
	\caption[Graph structure of basic geometric constraints]{The graph structure of three basic geometric constraints. The grey shaded region means the two points represent a line. }
	\label{fig:c1}
\end{figure}

\subsection{Task generalization by structural projection}

Given a graph structure $\mathcal{G}=\{V, E\}$, where $E$ is pre-defined to denote a geometric constraint type, projecting an image observation $o_{t} \in \mathbb{R}^{w  \times h \times c}$ on $\mathcal{G}$ means selecting task-relevant image features $\{f_{i}\}$ to fill in the graph nodes $V$. As shown in Fig. 1B, each image feature $\{f_{i}\}$ has two parts, $x_{i}$ and $y_{i}$, where $x_{i}$ is the feature descriptor\footnote{$x_{i}$ can be a hand-crafted one (such as ORB~\cite{rublee2011orb}) or in this paper, an image patch encoded by a convolutional neural network which is learnable.} and $y_{i}$ the feature's local coordinates. Suppose a graph neural network $g$ maps a graph $\mathcal{G}$ instance with nodes from $\mathbf{x}=\{x_{i}\}$ to a latent vector $\mathbb{z}_{t} \in \mathbb{R}^{d}$, we have,
\begin{equation}
\begin{aligned}
\mathbf{z}_{t}=g(\mathbf{x}|\mathcal{G})
\end{aligned}
\label{eq:sadf}
\end{equation}
Furthermore, assuming a function $\phi$ takes input of feature coordinates $\mathbf{y}=\{y_{i}\}$ and computes the related geometric constraint's error vector $\mathcal{E}_{t}$, then,
\begin{equation}
\begin{aligned}
\mathcal{E}_{t}=\psi(\mathbf{y}|\mathcal{G})
\end{aligned}
\label{eq:223}
\end{equation}

To summarize, $\mathbf{z}_{t}$ is the representation of a geometric constraint that defines the task and $\mathcal{E}_{t}$ is the adjoint geometric errors that are used in a controller (such as visual servoing~\cite{chaumette1994visual} or uncalibrated visual servoing~\cite{jagersand1995visual}.

Then, given two arbitrary images $o^{i}$ and $o^{j}$ showing a task scene with two different objects i and j respectively, their task representations $\mathbf{z}^{i}$ and $\mathbf{z}^{j}$ build a correspondence link by maximizing a similarity metric between them since they define the same task. We use a cosine distance as the similarity metric to normalize their magnitudes: 
\begin{equation}
sim(\mathbf{z}^{i}, \mathbf{z}^{j})= \frac{ \mathbf{z}^{i} \cdot \mathbf{z}^{j} }{ \norm {\mathbf{z}^{i}} \norm{\mathbf{z}^{j}}}
\label{asfd}
\end{equation}

Next, we will show that the similarity metric can be used to optimize a task function to learn generalizable task representations.

\subsection{Task function for generalizable representation learning}
In Eq. (1), a graph neural network $g$ can read out a graph instance but can not select task-relevant features from $\mathbf{x}$ to fill in the graph nodes. Built on top of $g$, a task function $T_{k}$, as introduced in~\cite{jin2020geometric}, is designed to select task-relevant image features to construct a graph that defines the task in a geometric constraint. $k$ denotes a geometric constraint type, such as a line-to-line constraint.
\begin{figure}[h]
\setlength{\belowcaptionskip}{-10pt}
	\includegraphics[width=0.45 \textwidth]{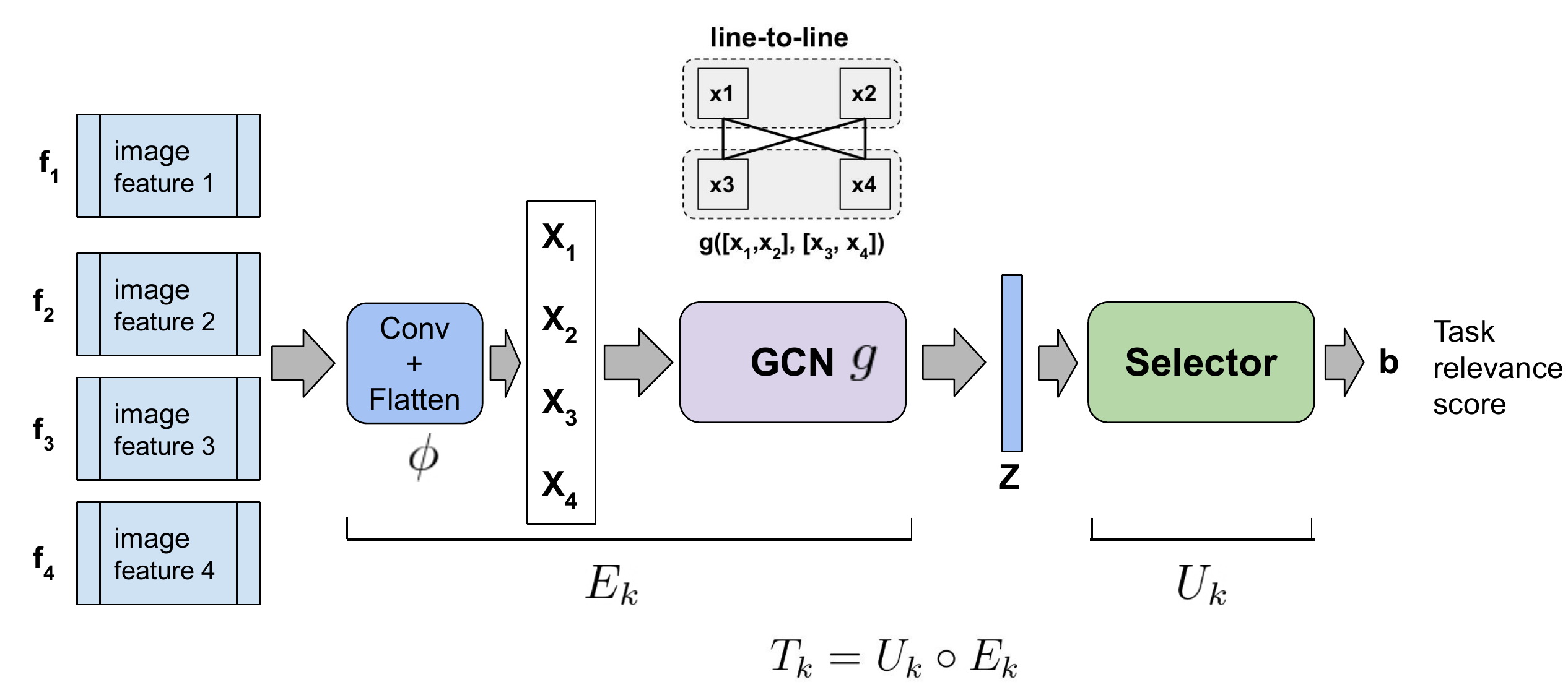}
	\centering
	\caption[Graph structured task function T.]{Graph structured task function $T_{k}$. } 
	\label{fig:t5}
\end{figure}

A task function $T_{k}$ has two parts (Fig. 5), a graph neural network-based encoder $E_{k}$ that reads out a graph instance as vector $\mathbf{z}_{t}$ and a selector $U_{k}$ that ranks the task-relevance score of a graph instance and selects out top $p$ graphs with high task-relevance scores. Similar to~\cite{jin2020visual}, we use a message-passing graph neural network~\cite{gilmer2017neural} with GRU updates as $g$ to encode the geometric constraints.

\textbf{Learning to select task-relevant geometric constraints:}
Given human demonstration videos, the above task function is optimized using VGS-IL (visual geometric skill-imitation learning~\cite{jin2020geometric}) by assuming the \textit{temporal-frame-orders} in demonstrated image frame sequence that defines the task. The underlying assumption is that demonstrated image transitions encode the task definition. So that the task function is optimized to encode the task definition by maximizing the probability of observed image state transitions in the demonstrated video frames. VGS-IL enforces $T_{k}$ learning to select task-relevant geometric constraints from an image. Due to page limits, readers can refer to~\cite{jin2020geometric} for more details. 

\textbf{Learning to select generalizable geometric constraints:}
To make the task function $T_{k}$ also select generalizable geometric constraints on different objects, we propose to combine VGS-IL with the following objective function in optimization:
\begin{equation}
\mathcal{L}_{sim}=\argmax_{T_{k}} \frac {1}{N} \sum sim(\mathbf{z}^{i}, \mathbf{z}^{j})
\label{sim_loss}
\end{equation}
where $N$ is the total number of image pairs from object i and j. Then by maximizing the objective function, we can enforce the task function $T_{k}$ learning to select similar geometric constraints on object i and j. 

\subsection{Categorical object generalizable VGS-IL}
Therefore, by joint optimization of VGS-IL~\cite{jin2020geometric} and the objective function in Eq. \ref{sim_loss}, we can enforce the task function $T_{k}$ selecting both task-relevant and generalizable geometric constraints, thus forming a generalizable representation of the task.

Based on the above idea, we propose CoVGS-IL (Categorical object generalizable VGS-IL), which takes in a set of human demonstration videos $\mathcal{S}=\{\{o^{i}_{t}\}, \{o^{j}_{t}\},...\}$ with categorical objects $\{i, j, ...\}$ and learns a task function $T_{k}$ that extracts task-relevant features on different objects with the same task functionality. In the joint optimization combing VGS-IL and the similarity loss in Eq. \ref{sim_loss}, a momentum update is applied to improve learning convergence. Algorithm details are concluded in Algorithm 1.


\begingroup
\removelatexerror
{\small
	\begin{algorithm}[h]
		\small
		\SetAlgoLined
		\KwIn{Human demonstration videos $\mathcal{S}=\{\{o^{i}_{t}\}, \{o^{j}_{t}\},...\}$ with categorical objects $\{i, j, ...\}$; demonstrator's confidence level $\alpha$ as shown in~\cite{jin2019robot}; graph structural priori $\mathcal{G}_{k}$ of a geometric constraint type $k$.}
		\KwResult{Optimal weights $\boldsymbol{\theta} ^{*}$ of the task function $T_{k}$}
		{Randomly initialize ${\theta}$ of $T_{k}$}, cloning $T_{k}$ as $T^{sim}_{k}$ with weights ${\theta_{sim}}$ \\
		Prepare dataset: define $\mathcal{D}_{s}=\{\}$, $\mathcal{D}_{img}=\{\}$ \\
		\For{each video $\{o^{i}_{t}\} \in \mathcal{S}$ }
		{
			Prepare state change samples $\mathcal{D}^{i}$ \\
			$\mathcal{D}_{s} \leftarrow $  Append $\mathcal{D}^{i}$
		}
		\For{each image frame $o_{t}$ in all demo videos $\mathcal{S}$}
		{
			Extracting feature points on $o_{t}$ using an off-the-shelf feature detector. \\
			$s_{t} \leftarrow $ Construct all graph instances by feature association according to $k$\\
			$\mathcal{D}_{img} \leftarrow $ Append $s_{t}$ \\
		}
		
		Shuffle $\mathcal{D}, \mathcal{D}_{img}$\\
		\For{n=1:N}{
			{Optimize $T_{k}$ using the ``temporal-frame-orders'' loss $\mathcal{L}$ for N1 steps:}
			$\theta ^{n+1} = \text{VGS-IL}(\mathcal{D}, \alpha, T_{k}, \theta ^{n})$\\
			Copy weights $\theta ^{n+1}$ to $\theta_{sim} ^{n}$. \\
			Optimize $T^{sim}_{k}$ using the similarity loss $\mathcal{L}_{sim}$ for N2 steps:\\
			$\theta_{sim} ^{n+1} = \theta_{sim} ^{n} + \nabla_{\theta_{sim}}\mathcal{L}_{sim}(\mathcal{D}_{img})$\\
			Perform one step momentum update on $T_{k}$:\\
			$\theta ^{n+1} = \beta \theta ^{n+1} + (1-\beta)\theta_{sim} ^{n+1} $, where $\beta \in (0,1)$ is a momentum coefficient. 
		}
		\caption{CoVGS-IL}
	\end{algorithm}
}
\endgroup

\subsection{Linkage to a robot controller}
The learned task function $T_{k}$ is then transferred to the robot in a feedback control loop (as shown in Fig. 2D) using uncalibrated visual servoing (UVS~\cite{jagersand1995visual}). Given an image observation $o_{t}$ at time t, the task function $T_{k}$ selects geometric constraints with adjoint geometric error output $\mathcal{E}_{t}$ (eq. 2) that are used in an UVS controller. Examples of computing $\mathcal{E}_{t}$ based on geometric constraint types are illustrated in Fig. 2. 

The uncalibrated visual servoing controller (UVS~\cite{jagersand1995visual}) uses online trial-and-error to directly estimates the Jacobian $\tilde {\mathbf{J}} \in \mathbb{R} ^{d \times n}$~\cite{Jagersand1997,Ramirez2016} which maps image observations to robot actions, where $d$ is the visual feature dimension, and $n$ is the number of joints used for control. As a result, given the output geometric error signal $\mathcal{E}_{t}$ at time t, the control law is formulated as:
\begin{equation}
\dot{\mathbf{q}}=-\lambda \tilde {\mathbf{J}}_{t}^{+} \mathcal{E}_{t}
\end{equation}
, where $\tilde {\mathbf{J}}_{t}^{+} $ is the pseudo-inverse of $\tilde {\mathbf{J}}_{t}$.

A UVS controller starts with an initial estimation of $\tilde {\mathbf{J}}_{0}$ by exploratory motions, which measure how robot action affects the numerical value changes of observed features. The initial Jacobian estimation via trial-and-errors is based on the following equation:
\begin{equation}
\tilde {\mathbf{J}}_{0} = \left [ \left [\frac{\Delta \mathbf{e}_{q_{1}}}{\Delta q_{1}}\right ] ... \left [\frac{\Delta \mathbf{e}_{q_{m}}}{\Delta q_{m}}\right ]\right ]
\end{equation}
, which means the $\Delta q_{i}$ amount of joint i movement results in geometric error change $\Delta \mathbf{e}_{q_{i}}$. After initial estimation, this Jacobian is then continuously updating online via Broyden update during iterations:

\begin{equation}
\tilde {\mathbf{J}}_{t+1}=\tilde {\mathbf{J}}_{t}+\alpha \frac{(\Delta \mathbf{e} - \hat{J}_{t}\Delta q)\Delta q^{T}}{\Delta q^{T}\Delta q}
\end{equation}
, where $\alpha$ is the update step size.


\section{Evaluation}
\textbf{Evaluation objectives:}
We perform evaluations on two objectives. (1) Can CoVGS-IL learn generalizable task representations for categorical objects? (2) From an imitation learning system perspective, can CoVGS-IL enable a real-world robot to learn generalizable skills by watching human demonstration videos?

\textbf{Experimental setup:}
Our experimental setup is shown in Fig. 6A where an eye-to-hand camera is used to record human demonstrations and guide the robot actions. We use a hammering task (Fig. 6B and C), which requires a robot holding the hammer and hit the nail with a proper orientation\footnote{Striking motion in the final steps is not considered since it involves force/torque control which is beyond our study scope as discussed in Section V Conclusion}. The hammer is grasped with a random pose at each run. 

\begin{figure}[h]
\setlength{\belowcaptionskip}{-10pt}
	\includegraphics[width=0.5\textwidth]{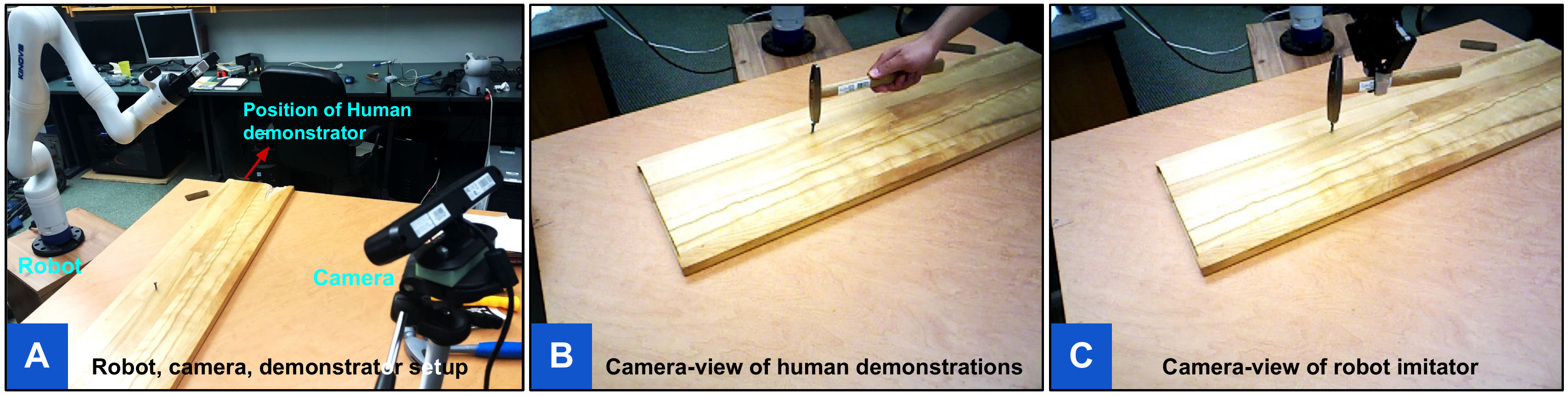}
	\centering
	\caption[Task design for categorical object generalization evaluation.]{\footnotesize Experimental setups. \textbf{A}: Robot-camera-demonstrator setup. \textbf{B}: Human demonstrates the hammering task from the camera's view. \textbf{C}: Robot imitates the hammering task from the camera's view. }
	\label{fig:6_experiment_setup}
\end{figure}

From a geometric perspective, the task can be specified as a point-to-point constraint (a point on hammer bottom to a point on nail top) and line-to-line parallelism (an edge on the hammer part to the body of the nail). The learned task function is required to extract correct geometric constraints that specify the task while considering generalization to different hammers.

\subsection{Evaluation of generalizable task representation}
\textbf{Training:}
As shown in Fig. 7, we recorded 30 human demonstration videos using three hammers (A, B and C) with ten videos for each hammer. Each video lasts about 3 to 8 seconds with about 60 to 180 frames. An ORB feature detector~\cite{rublee2011orb} is used to extract feature points on an image $o_{t}$. Then we run CoVGS-IL to train a graph-structured task function as described in Algorithm 1.

\begin{figure}[h]
\setlength{\belowcaptionskip}{-10pt}
	\includegraphics[width=0.45\textwidth]{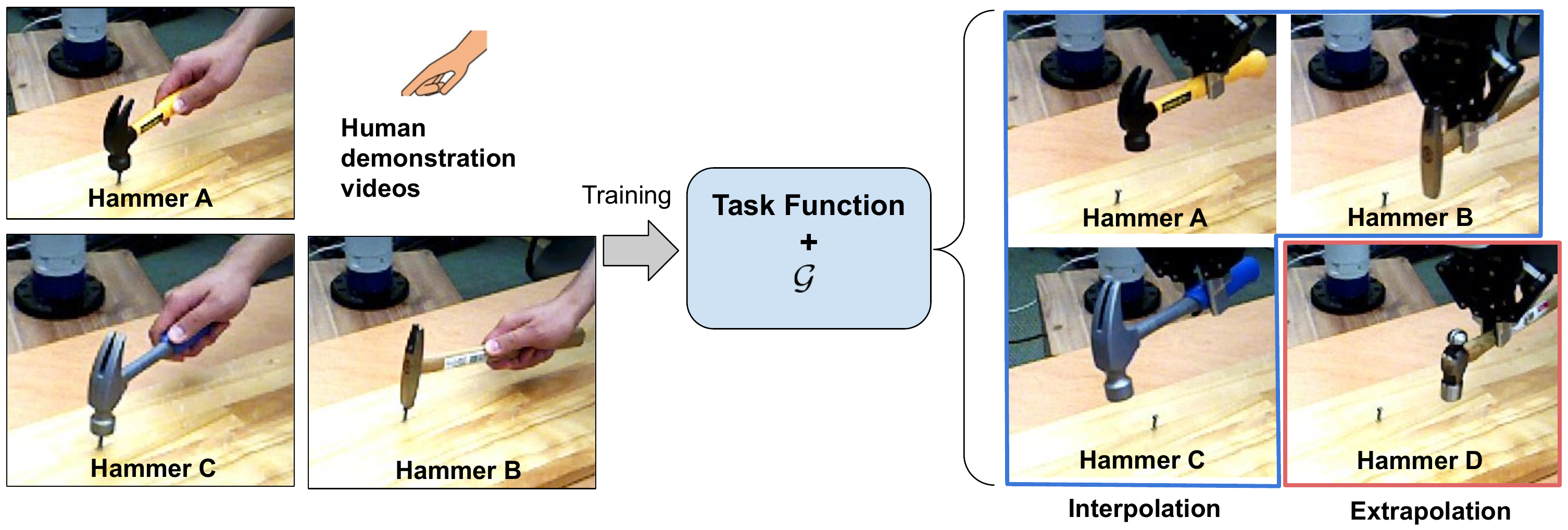}
	\centering
	\caption[Categorical object generalization evaluations.]{\footnotesize Evaluating categorical object generalization regarding interpolation and extrapolation.}
	\label{fig:6_categorical_object_gen}
\end{figure}

\textbf{Evaluation protocol:}
Our testing strategy is to check if the learned task function will select out correct and consistent geometric constraints using five robot videos recorded by humans moving the robot with random motions, and then test the task execution performance that will be elaborated in Section \ref{secttt}.

\textbf{Metrics:}
For video evaluation, we design two metrics. As described in~\cite{jin2020geometric}, (1) $Acc$ to measure selection accuracy by calculating the percentage of image frames with correct geometric constraint selection and (2) $ConAcc$ to measure selection consistency by an auto-correlation measurement over the time-series geometric error output $\{\mathcal{E}_{t}\}$.

\textbf{Baseline:}
Since there are no existing methods that learn generalizable geometric constraints from human demonstration videos, we hand-designed a \textit{hand-tracking-baseline} based on a video-tracking module as defined in~\cite{jin2020geometric}. Moreover, since the \textit{hand-tracking-baseline} does not generalize to different hammers, we manually specify the features to be tracked for each hammer. As a result, the hand-tracking baseline is a strong one since the above special considerations.

\begin{table}[h]
	\scriptsize
	\centering
	\begin{tabular}{@{}cccccc@{}}
		\toprule
		\multicolumn{2}{c}{Methods}    & \multicolumn{2}{c}{Hand-tracking} & \multicolumn{2}{c}{Ours}        \\ \midrule
		\multicolumn{2}{c}{Tasks}      & Acc               & ConACC        & Acc              & ConAcc       \\ \midrule
		\multirow{2}{*}{A} & PP & 100.0\% $\pm$ 0.0\%     & 1.00 $\pm$ 0.00  & 100.0\% $\pm$ 0.0\%   & 0.91 $\pm$ 0.05 \\
		& LL & 100.0\% $\pm$ 0.0\%  & 1.00 $\pm$ 0.00  & 95.4\% $\pm$ 3.1\%  & 0.89 $\pm$ 0.09 \\ \cline{1-6} 
		\multirow{2}{*}{B} & PP & 89.4\% $\pm$ 8.7\%   & 0.89 $\pm$ 0.07  & 88.4\% $\pm$ 7.1\% & 0.80 $\pm$ 0.11 \\
		& LL & 81.4\% $\pm$ 17.7\%  & 0.49 $\pm$ 0.59  & 86.2\% $\pm$ 7.2\%  & 0.60 $\pm$ 0.28 \\ \cline{1-6} 
		\multirow{2}{*}{C} & PP & 100.0\% $\pm$ 0.0\%   & 1.00 $\pm$ 0.00  & 97.2\% $\pm$ 3.5\%  & 0.60 $\pm$ 0.28 \\
		& LL & 96.0\% $\pm$ 4.1\%     & 0.88 $\pm$ 0.13  & 98.6\% $\pm$ 2.3\%  & 0.89 $\pm$ 0.08 \\ \bottomrule
	\end{tabular}
	\caption[Categorical object generalization regarding interpolation.]{\footnotesize Evaluation on categorical object generalization using hammer A, B and C. For convenience, ``PP'' refers to ``point-to-point'' constraint and ``LL'' as ``line-to-line'' constraint.}
	\label{table:6_interpolation_of_hammers}
\end{table}

\textbf{Categorical object generalization results:}
Following the above guidelines, we report the evaluation results of the task function's performance in hammer A, B, and C. Results (Table \ref{table:6_interpolation_of_hammers}) show the learned task function's performance matches the \textit{hand-tracking} baseline.

\begin{table}[h]
	\scriptsize
	\centering
	\begin{tabular}{@{}cccccc@{}}
		\toprule
		\multicolumn{2}{c}{Methods}    & \multicolumn{2}{c}{Hand-tracking} & \multicolumn{2}{c}{Ours}       \\ \midrule
		\multicolumn{2}{c}{Tasks}      & Acc               & ConACC        & Acc             & ConAcc       \\ \midrule
		\multirow{2}{*}{D}  & PP & 88.3\% $\pm$ 8.9\%   & 0.71 $\pm$ 0.15  & 91.1\% $\pm$ 4.7\% & 0.87 $\pm$ 0.08 \\
		& LL & 84.0\% $\pm$ 15.1\%  & 0.61 $\pm$ 0.42  & 86.7\% $\pm$ 7.0\% & 0.74 $\pm$ 0.21 \\ \bottomrule
	\end{tabular}
	\caption[Categorical object generalization regarding extrapolation.]{\footnotesize Categorical object generalization regarding extrapolation using hammer D.}
	\label{table:6_extrapolation_of_screwdriver_hammer}
\end{table}

\begin{figure*}[h]
\setlength{\belowcaptionskip}{-10pt}
	\includegraphics[width=0.7\textwidth]{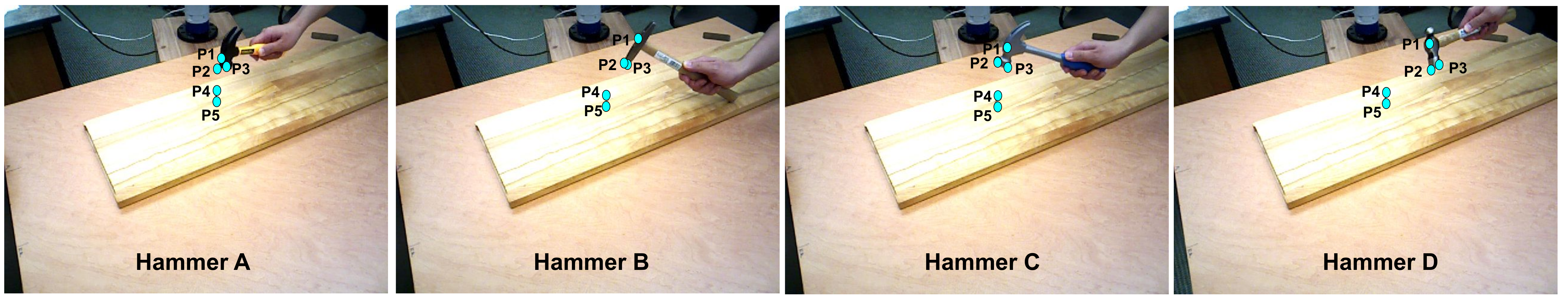}
	\centering
	\caption[Visualization of the select-out geometric constraints in the four types of hammers.]{ \footnotesize Visualization of the select-out geometric constraints in the four hammers. The learned task function selects out image features (green dots) that share common characteristics among different hammers. On each hammer, the hammering task is defined by a line-to-line constraint (two lines defined by P1, P2, P4 and P5) and a point-to-point constraint (two points P3 and P4). Task specification correspondence is built by selecting the same geometric constraints (a line-to-line and a point-to-point) that define the same task.}
	\label{fig:4hammers_selection}
\end{figure*}

In addition, we are also interested in testing hammer D, which is never used during training. We wonder if the learned task function will also generalize to hammer D since it shares a similar shape with hammer A and C, and a similar texture with hammer B. The testing results (Table \ref{table:6_extrapolation_of_screwdriver_hammer}) surprisingly show that learned task function's performance matches the \textit{hand-tracking} baseline moderately. Again, note that the \textit{hand tracking} baseline does not generalize to hammer D since it is hand designed.

\textbf{Visualize task specification correspondence:}
Then, we visualize the select-out geometric constraints in the four types of hammers (Fig. \ref{fig:4hammers_selection}). Next, we empirically evaluate if the task function learns to build \textit{task specification correspondence} by visualizing the learned task representation for the four hammers. We visualize the output vector $\mathbf{z}_{t}$ for each hammer in 16 time steps of an evaluation video. The learned task function maps a point-to-point constraint (in the form of a graph) to a representation vector $\mathbf{z}_{t}$. For simplicity, only the first element of vector $\mathbf{z}_{t}$ is visualized. The color map shows how it changes along time steps (horizontal axis) and across different hammers (vertical axis).
\begin{figure}[h]
\setlength{\belowcaptionskip}{-10pt}
	\includegraphics[width=0.5\textwidth]{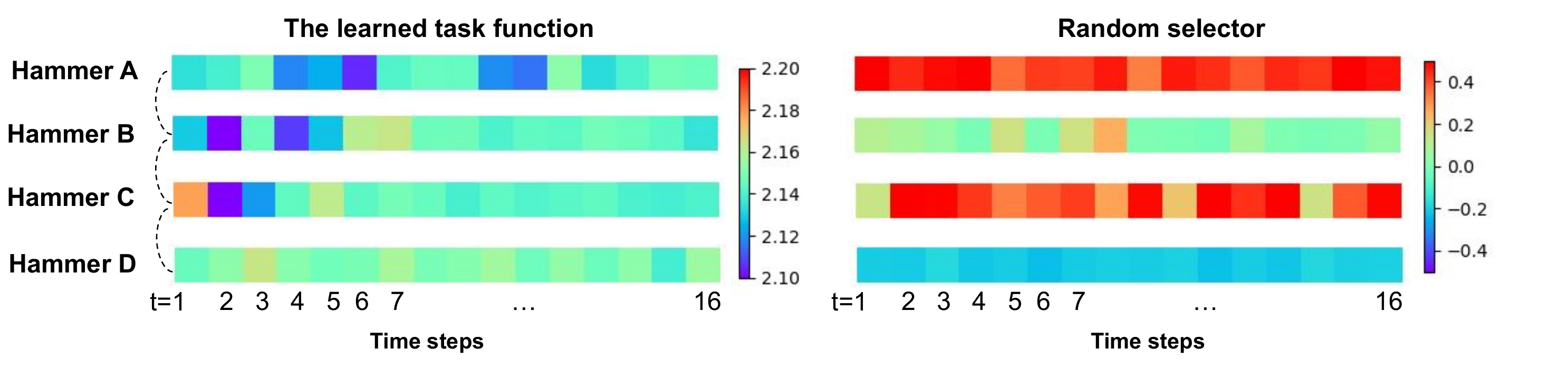}
	\centering
	\caption[Visualize the task-specification correspondence]{\footnotesize Task specification correspondence visualization of the point-to-point constraint which involves two points coincidence to define the task.  \textbf{Left:} the learned task function’s output $\mathbf{z}_{t}$. \textbf{Right:} a random selector's output $\mathbf{z}_{t}$. A well trained task function builds the task specification correspondence by mapping graph instances of different features points on the four hammers to the same task representation $\mathbf{z}_{t}$. It will also outputs a consistent task representation $\mathbf{z}_{t}$ along different time steps since each image frame also defines the same task using relevant geometric constraints. 
	}
	\label{fig:6_task_embedding_the_same}
\end{figure}

We compare our method to a random selector that randomly selects a geometric constraint and input the selection to the learned task function to output a vector for visualization. Results in Fig. 9 show that the task representation $z_{t}$ for hammer A, B, C and D are similar to each other while allowing slight value changes.  In contrast, the random selector outputs vectors of the four hammers that do not match each other.

\subsection{Evaluation of the imitation learning system on the robot}
\label{secttt}
Lastly, we transfer the learned task function to the robot to guide the UVS controller forming a full imitation learning system.

\textbf{Evaluation protocol:}
 For hammers A, B, C and D, each hammer is evaluated using 10 robot trials with each trial changing the grasping pose. A successful trial is defined as the robot holding the hammer and hit the nail without considering knocking it down, which requires extending our method to consider force/torque control. The success rate is reported as the evaluation metric.
 
\textbf{Baselines:}
We design three baselines for comparison. (1) \textit{Behavior cloning~\cite{Argall2009}}: We collect images and robot joint reading pairs ($o_{t}, q$) for hammer A, B and C. Then, we supervised training a four-layer convolutional neural network that regresses a function mapping from image to robot joint values and uses the trained model on the robot. Each hammer has ten human demonstrations via kinesthetic teaching. (2) \textit{Trajectory replay}: we use a pre-recorded trajectory for each hammer by human kinesthetic teaching. In testing, we replay the recorded trajectory. (3) \textit{RL (SAC)}~\cite{haarnoja2018soft}: we train a soft-actor-critic (SAC) agent for each hammer. The policy maps image observation $o_{t}$, to the robot’s 7 joint velocities $\dot{q}$. Since there is no ground truth hammer pose, we use a reward classifier, similar to ~\cite{dulac2019challenges} as the reward function.

\textbf{Skill generalization results:}
Results are shown in Table \ref{table:8_all_results}, which indicates our method outperforms the baselines. The \textit{behaviour cloning} baseline is hard to precisely hit the nail, which is highly dependent on the learned regressor's accuracy. The \textit{Trajectory replay} baseline only works for the first trial but fails during successive trials when the robot randomly grasps the hammer. The \textit{RL (SAC)}~\cite{haarnoja2018soft} baseline failed in training after about 7 hours running on the robot.

\begin{table}[h]
\footnotesize
\centering
\begin{tabular}{lcccc}
\hline
\multirow{2}{*}{Methods}       & \multicolumn{4}{c}{Categorical object generalization} \\ \cline{2-5} 
                               & hammer A       & B          & C           & D         \\ \hline
Behavior cloning                             & 20\%           & 20\%       & F           & F         \\
Trajectory replay                             & 10\%           & 10\%       & 10\%        & 10\%      \\
RL (SAC)                       & F            & F        & F         & F       \\

\textbf{Ours} & 100\%          & 90\%       & 100\%       & 70\%      \\ \hline
\end{tabular}

\caption[Overview of results]{\footnotesize Results in the hammering task considering skill generalization. {``F''} marks failure.}
\label{table:8_all_results}
\end{table}

\section{Conclusion}
This paper takes a geometric perspective on the problem of generalizable task learning using human demonstration videos. Three insights enable the success of our method: (1) the insight that a geometric constraint is a multi-entity relationship between a set of image features, and the relationship can be represented using relation encoders (graph neural network in this paper); (2) the insight from visual servoing literature~\cite{chaumette1994visual} that a wide range of robotic manipulation tasks can be specified using geometric constraints; (3) the insight that task specification is consistent under different task settings (\textit{task specification correspondence}), which gives a clue for generalizable task representation learning. 

One main limitation here is the huge computational cost required when the number of handcrafted feature points increases since the selector $U_{k}$ is trained by constructing all possible graph instances that requires a combinatorial search. This limitation is more obvious when using a graph to represent complex geometric constraints like the point-to-conic. Further investigation on learning task-relevant feature detector~\cite{kulkarni2019unsupervised} instead of a handcrafted one will solve the problem. Besides, exploring combing visual inputs with force feedbacks~\cite{jin2020offline} from a geometric perspective will improve the applicability of our method in more manipulation tasks.

	\addtolength{\textheight}{-0 cm}   
	



	
	
	\bibliographystyle{IEEEtran}
	\bibliography{IEEEabrv,IEEEexample}
	
\end{document}